\documentclass[AMS,STIX1COL]{WileyNJD-v2}
\usepackage{xcolor}
\usepackage{soul}
\usepackage{comment}
\usepackage{subcaption}
\usepackage{caption}
\articletype{Article Type}%

\begin{document}

\title{Automating Document Classification with Distant Supervision to Increase the Efficiency of Systematic Reviews}


\author[1]{Xiaoxiao Li}
\author[2]{Rabah Al-Zaidy}
\author[1]{Amy Zhang}
\author[3]{Stefan Baral}
\author[1]{Le Bao*}
\author[4]{C. Lee Giles}


\address[1]{\orgdiv{Department of Statistics}, \orgname{Pennsylvania State University}, \orgaddress{\state{Pennsylvania}, \country{USA}}}

\address[2]{\orgdiv{Department of Information and Computer Science},
\orgname{King Fahad University of Petroleum and Minerals},
\country{Saudi Arabia}}

\address[3]{\orgdiv{Bloomberg School of Public Health}, \orgname{John Hopkins University}, \orgaddress{\state{Maryland}, \country{USA}}}

\address[4]{\orgdiv{Information Sciences and Technology}, \orgname{Pennsylvania State University}, \orgaddress{\state{Pennsylvania}, \country{USA}}}

\corres{*Le Bao. \email{lebao@psu.edu}}

\presentaddress{Department of Statistics\\
Penn State University\\
514C Wartick Lab\\
University Park, PA 16802}

\abstract[Abstract]{
Objective: Systematic reviews of scholarly documents often provide complete and exhaustive summaries of literature relevant to a research question. However, well-done systematic reviews are expensive, time-demanding, and labor-intensive. Here, we propose an automatic document classification approach to significantly reduce the effort in reviewing documents.
Methods: We first describe a manual document classification procedure that is used to curate a pertinent training dataset and then propose three classifiers: a keyword-guided method, a cluster analysis-based refined method, and a random forest approach that utilizes a large set of feature tokens. As an example, this approach is used to identify documents studying female sex workers that are assumed to contain content relevant to either HIV or violence. We compare the performance of the three classifiers by cross-validation and conduct a sensitivity analysis on the portion of data utilized in training the model.
Results: The random forest approach provides the highest area under the curve (AUC) for both receiver operating characteristic (ROC) and precision/recall (PR). Analyses of precision and recall suggest that random forest could facilitate manually reviewing 20\% of the articles while containing 80\% of the relevant cases. Finally, we found a good classifier could be obtained by using a relatively small training sample size.
Conclusions: In sum, the automated procedure of document classification presented here could improve both the precision and efficiency of systematic reviews, as well as facilitating live reviews, where reviews are updated regularly.}


\keywords{Text mining, Literature screening, Classification}

 \maketitle


\section{Introduction}\label{sec:intro}
Systematic literature review (SR) has been a key step of evidence-based healthcare research, which distinguishes itself from any ad hoc literature selection by its explicit and systematic approach. Studies for specific diseases or interventions are collected, summarized, and extensively reported via SR to aid decision making by physicians, policy makers, and patients. Examples include the global burden of disease attributable to mental and substance use disorders \cite{whiteford2013global}, the long-term health consequences of child abuse \cite{norman2012long}, the effect of antibiotic prescribing on antimicrobial resistance \cite{costelloe2010effect}, and so on. As such, guidelines for conducting and reporting SRs have been developed to improve the quality of SRs and increase their transparency. One notable example is the Preferred Reporting Items for Systematic Reviews and Meta-Analyses (PRISMA), which consists of a 27-item checklist and four-phase diagram to help improve the reporting of systematic reviews \cite{moher2009preferred}.  \citet{khan2003five} defined five steps in forming a SR study, which we summarize here:
\begin{enumerate}
    \item Framing clear and unambiguous questions to guide the review. These are used to define review protocol and criteria for inclusion or exclusion.
    \item Identifying relevant work through an exhaustive search of the literature. Selection criteria are based on the review questions defined previously.
    \item Assessing the quality of studies using, for example, design-based quality checklists. 
    \item Summarizing the evidence through tabulation of study characteristics, quality, and effects as well as use of statistical methods for meta-analysis.
    \item Interpreting the findings. Explore the risk of publication or other biases and determine the quality of the summary.
\end{enumerate}
To complete high quality systematic reviews, significant efforts are invested into developing and implementing appropriate search strategies including searching through tens of thousands of research articles to include in a SR. Automatic document classification (ADC) can help distinguish relevant work from others and greatly improve the efficiency of an SR project. ADC has been an important research topic for years and many techniques have been developed and applied in this field, such as the Na\"ive Bayes classifier ~\cite{agrawal2000}, support vector machines ~\cite{joachims1998}, and recent deep neural network methods ~\cite{collobert2008}. Using such machine learning algorithms to assist the process of SR, we hope researchers' workload for reading thousands of documents could be greatly alleviated.  

In both the Global Fund 2017-2021 strategy and PEPFAR 3.0, the need for empirical data-driven responses have been highlighted as central to informing an effective HIV response. In response, there have been significant efforts to build data repositories for particular populations disproportionately affected by HIV, generally called key populations.  Critical metrics to inform decision making include population size, HIV prevalence and incidence data, treatment and prevention cascade indicators, violence indicators, and consistent condom use indicators. Under the PRISMA guidelines, we systematically reviewed all published data related to female sex workers, and assembled data from studies published in all low and middle-income countries including all Sub-Saharan African countries except for the Seychelles. To compare with a traditional systematic review, we used ADC to identify journal articles that characterize key features of female sex workers as our baseline.

The rest of the article is organized as follows. In Section \ref{sec:data}, we describe the document labelling procedure and the text data extraction procedure. In Section \ref{sec:method}, we propose three approaches for automatically classifying documents into the relevant class versus the irrelevant class. In Section \ref{sec:result}, we compare the performance of three approaches and conduct a sensitivity analysis on the portion of data utilized in training the model. Section \ref{sec:conc} concludes with discussions.
 

\section{Data}\label{sec:method}

We describe the procedure of manually labeling the status of articles (relevant versus irrelevant) in Section 3.1 and the natural language processing (NLP) tools for preprocessing the text data in Section 3.2.

\subsection{Manual Document Classification}\label{sec:classification}

Manual document classification has been a reliable process to select relevant documents for specific research purposes. The research team at John Hopkins University manually reviewed journal articles from PubMed, EMBASE, Global Health, SCOPUS, PsycINFO, Sociological Abstracts, CINAHL (Cumulative Index to Nursing and Allied Health Literature), Web of Science and POPLine. The reviewed articles were labeled as relevant if they included either HIV data or violence data among female sex workers and labeled as irrelevant otherwise. All the articles and reports used in the study must have been either published in a peer-reviewed journal, presented as an abstract at a scientific conference, or available on the web from governmental or non-governmental sources between 2006 and 2017. Works published in languages other than English, and studies where the sample size was less than $50$ were not included. The review procedure is described below. 

Titles, abstracts, and citation information were screened by two members of the study team. Full-text articles were obtained after the initial screening. Title, abstract and full-text review were conducted independently by two reviewers using standardized data extraction forms in Covidence, a tool designed to help facilitate the systematic review process. Differences in data extraction were resolved through consensus and referral to a senior study team member when necessary. There were some documents in the database with no abstract information and were excluded during analysis. Following the review process above, we obtained $10,718$ documents manually labeled with relevant/irrelevant which could serve as ground truth. Among these vetted text documents, $1,171$ papers were marked as relevant and $9,547$ as irrelevant. 



\subsection{Data Preprocessing}\label{sec:data}
We combined the titles and abstracts of the 10,718 documents and normalized the raw text data as follows: 
\begin{description}
     \item[Tokenization:] We split the text into individual word tokens based on white space and segmented the text into basic linguistic units. All punctuation, numbers, special characters, and languages other than English were removed. 
     \item[Lemmatization:] Following Mechura's \cite{mechura2016data} English lemmatization list, we grouped together various derivative forms of a word which share similar semantic meaning such that they could be analyzed as a single item. For example, the tokens "am", "is" and "are" became the single token "be". 
    \item[Stemming:] We removed word suffixes and conflated the resulting morphemes with the Porter stemmer \cite{porter2006algorithm} which leads to a crude affix chopping. For example "automates" and "automation" all reduce to "automat" using the Porter stemmer.
\end{description}

The tokens obtained after pre-processing for each document were translated into a numerical vector representation called the document-term-matrix (DTM). A typical representation of the DTM is as a $D \times P$ matrix, where $D$ is the number of documents and $P$ the number of unique tokens after pre-processing. The $(d, i)^{th}$ entry in the DTM records the frequency of token $i$ in document $d$. Note that many tokens are common across documents but may not be relevant to the SR search criteria, such as articles, prepositions, and certain common verbs (e.g., "the", "and", "be"). Thus we scaled token frequency within a document by the inverse of a token's frequency across all documents, which is called term-frequency-inverse-document-frequency (TFIDF) \cite{jones1972statistical}:
     \begin{equation*}
         \text{tfidf}_{i,d} = \text{tf}_{i,d} \cdot \text{idf}_{i},
     \end{equation*}
where $\text{tf}_{i,d}$ is the term-frequency for token $i$ in document $d$ and $\text{idf}_{i}$ is the inverse document frequency for token $i$. For our study, we used the logarithmically scaled inverse fraction of documents containing token $i$: $\text{idf}_i = \text{log} \frac{N}{\text{df}(i)}$ where $N$ is the total number of documents and $\text{df}(i)$ is the frequency of token $i$ in documents. Using TF-IDF, tokens which are common across all documents are treated as less informative and thus less important. This measures how much information the word provides.  
 


\section{Document Classification Models} \label{sec:method}
We compared three classifiers that automatically identify articles of interest which are based on query searching, clustering, and a broadly used machine learning algorithm. These are described in detail in the following subsections.  

\subsection{Keyword-Guided Approach}\label{sec:method1}
Our goal was to identify documents with studies of female sex workers which included HIV or violence data. As a baseline model, we first examined document classification based on query searching. Table \ref{tab:keywords} lists the keywords in three categories: female sex workers (FSW), human immunodeficiency virus (HIV) and violence (Violence). To retrieve articles, key words across the three categories were combined in a query with the structure "FSW \texttt{AND} \texttt{(}HIV \texttt{OR} Violence\texttt{)}" under a Boolean search algorithm \cite{angione1975equivalence}.

\begin{center}
\begin{table}[!h]
\centering
\caption{Search Criteria Keywords.\label{tab:keywords}}%
\begin{tabular*}{500pt}{@{\extracolsep\fill}ll@{\extracolsep\fill}}
\toprule
\multicolumn{2}{c}{Female Sex Workers (\textbf{FSW})} \\
\midrule
MeSH & Prostitution, Sex Worker   \\
\midrule
TW  & prostitut*, commercial sex, transactional sex, sw, fsw, csw, sex trade, trade sex   \\
\toprule
\multicolumn{2}{c}{\textbf{HIV}} \\
\midrule
MeSH & HIV, acquired Immunodeficiency Syndrom, HIV Infections \\
\midrule
TW & human immunodeficiency virus*, acquired immunodeficiency syndrome*, HIV*, AIDS \\
\toprule
\multicolumn{2}{c}{\textbf{Violence}} \\
\midrule
MeSH & Violence, Domestic Violence, Workplace Violence, Crime Victims, Battered Women,Rape,
Homicide, Coercion \\
\midrule
TW & Violen*, crime*, offense*, abuse*,victim*,rape*,assault*,batter*,extort*,intimidat*,
exploit*,IPV,IPSV \\

\bottomrule
\end{tabular*}
\end{table}
\end{center}


\subsection{Cluster Refinement}\label{sec:method2}

The baseline model described in Section \ref{sec:method1} can be viewed as a tree-based classifier with three major predefined clusters of terms. We further divided those three large categories into finer clusters based on word roots, and allowed more flexible structures for classification. 

Constructing the finer clusters from the terms we extracted in Section ~\ref{sec:data} was a nontrivial process. Popular stemming methods truncate the ends of words, which often includes the removal of derivational affixes. Although the stemming process performs well in terms of reducing dimensionality, it also has the potential of creating ambiguity \cite{schofield2016comparing}. Due to the "crude chopping" of the stemming procedure, many outputs are not recognizable words. For example "stay" becomes "stai" using the Porter stemmer. Also, affixes are essential in meaning in English, but stemmers fail to capture this effect extensively; for example, "recondition" shares a stem but not the root meaning of "recondite". These issues may result in inaccurate stemming of words that share the same root but have different meanings, such as "absolutely" and "absolution". To mitigate this problem, we grouped together words of similar semantic meaning to guarantee that major words in the three categories can be represented by the same root. As a result, we partitioned the three major categories of tokens into the following 15 ``finer" clusters: "hiv", "fsw", "violence", "offense" "abuse", "torture", "rape", "victim", "assault", "harass", "extort", "homicide", "coercion", "ipv", "exploit". 

To obtain the vector representation of those 15 clusters, we created a document-cluster-matrix and calculated TF-IDF based on the combined frequencies of the terms in each cluster. Each document was represented by a 15-dimension vector of TF-IDFs. We then used random forest \cite{randomforest} to classify the documents with 15 features. Random forest uses a bagging ensemble method and decision trees constructed by  subset of data to provide more stable and accurate classification results. 

The two classes of documents were highly imbalanced: around one relevant document for every ten irrelevant ones. To relieve the distortion, we randomly down-sampled observations from the irrelevant class and kept similar sizes between the two classes when training each branch of the classification tree. We generated 500 classification trees, counted the votes for document $d$ being relevant, $Y_d=1$, among the 500 constructed trees, and used the proportion of votes as the fitted probability of $\hat{P}(Y_d=1)$.


\subsection{Top N Tokens Model}\label{sec:method3} 

In addition to the three major categories for tokens used in Section ~\ref{sec:method1} and Section ~\ref{sec:method2}, there still may have been some tokens that contained important information for classification purposes. Therefore, we introduced a screening method to identify other significant tokens and included these tokens as additional covariates in the classification model. 

For each token $i$, we used a two-sample t-test to determine whether its mean TF-IDF differed statistically between the relevant class and the irrelevant class. The test statistic was $\bar{x}_{i1} - \bar{x}_{i2}$ divided by the un-pooled variance, where $\bar{x}_{i1}$ is the mean TF-IDF value for token $i$ within the relevant documents and $\bar{x}_{i2}$ the mean TF-IDF value for token $i$ within the irrelevant documents. If a token $i$ was useful for distinguishing between the two groups/classes, the t-statistic was expected to have larger absolute value. We ranked tokens based on their absolute t-statistics and picked the top 20, 50, 100, 250 and 500 tokens as new features in addition to the 15 clusters defined in Section ~\ref{sec:method2} and constructed a set of models using these top N features.


\section{Results}\label{sec:result}



We refer to the three methods described in Section \ref{sec:method} as Model 1 (keyword-guided approach), Model 2 (random forest with 15 clusters) and Model 3 (random forest with 15 clusters and top N terms). We trained the three models using 5-fold cross validation and recorded their performance using the receiver operating characteristic (ROC) and precision and recall (PR). Models 2 and 3 produce probabilistic classifications, thus we could additionally compare ROC and PR curves by varying the cut-off probability for classifying document $d$ as relevant. For these models, we also reported the area under the curve (AUC) as a summary metric. 

\begin{figure}[!htb]
  \centering
\begin{subfigure}[b]{0.45\textwidth}
  \centering
  \caption{}
  \includegraphics[height=2.2in]{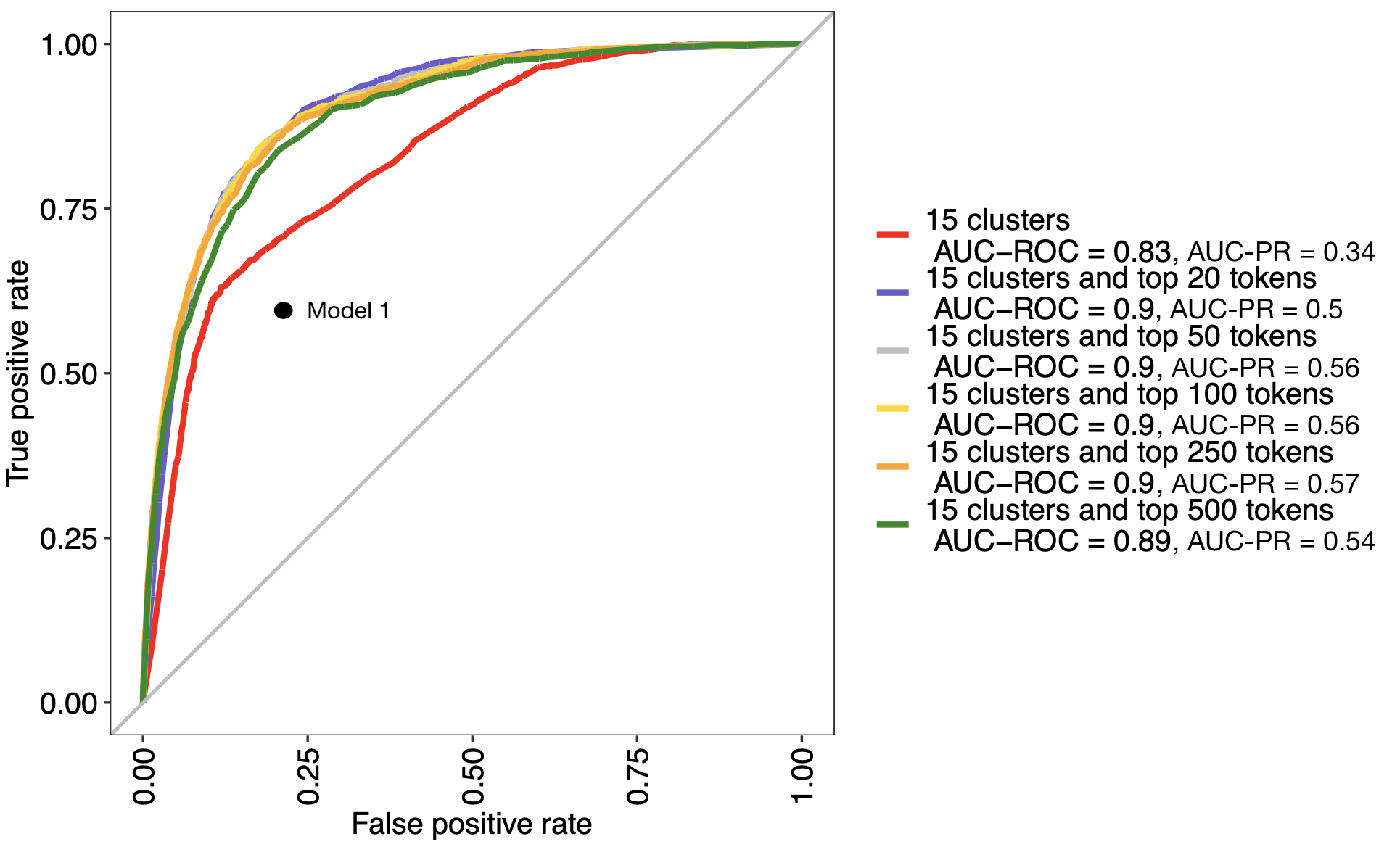}
\end{subfigure}
\begin{subfigure}[b]{0.45\textwidth}
  \centering
  \caption{}
  \includegraphics[height=2.2in]{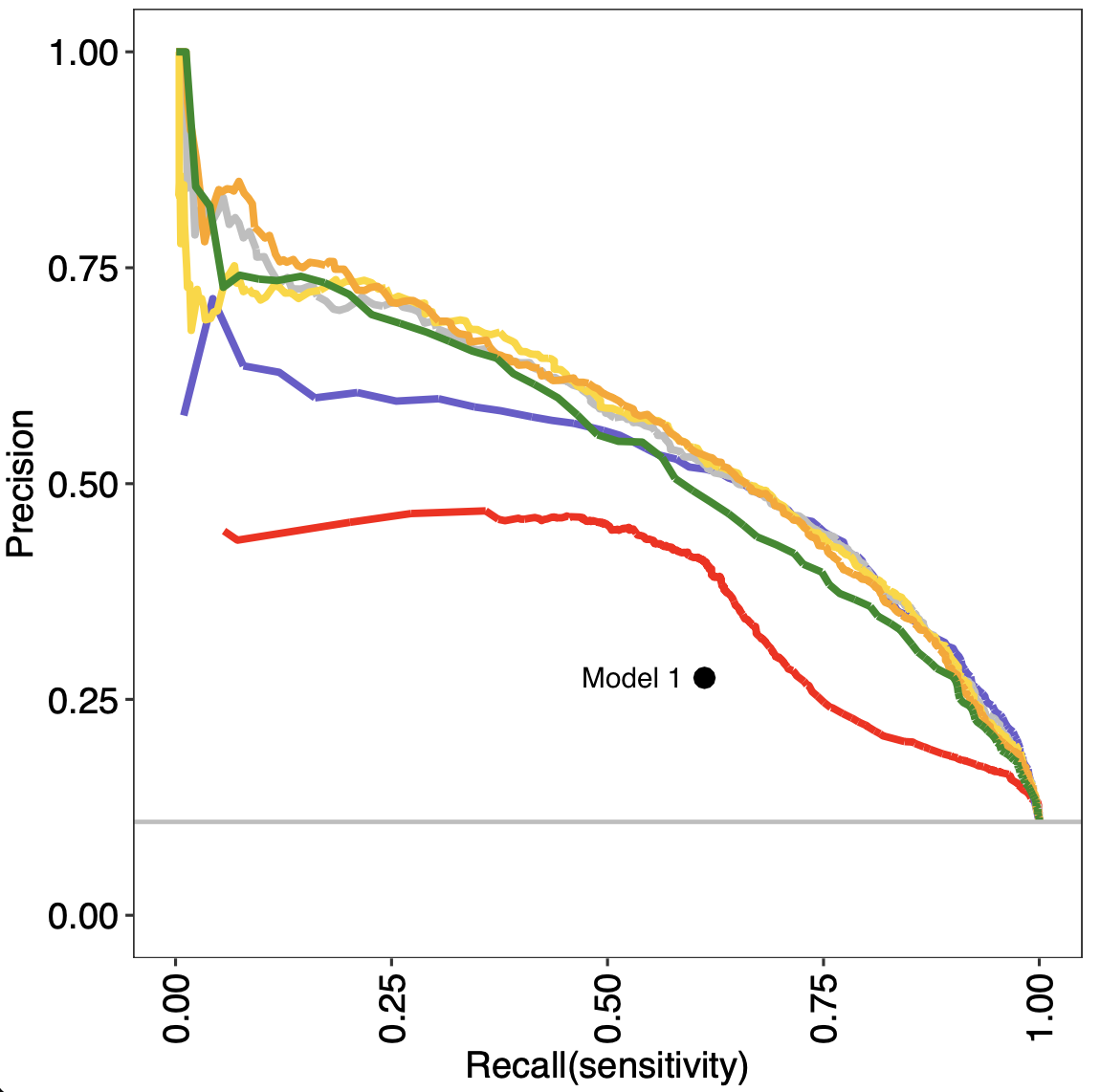} 
\end{subfigure}

\captionsetup{justification=centering,margin=2cm}
\caption{(a) ROC and (b) PR curves for Model 1 (keyword-guided approach), Model 2 (random forest with 15 clusters) and Model 3 (15 clusters with top 20, 50, 100, 250, and 500 tokens). As a summary metric, area under the curve (AUC) is provided in the legend.}
\label{fig:prec}
\end{figure}

Figure \ref{fig:prec}(a) shows the ROC curves of all models for comparison. All iterations of Model 3 outperform Model 2, which in turn outperforms the keyword-guided approach of Model 1. Varying the number of additional tokens (from $N=20$ to $N=500$) in Model 3 does not obviously change its performance. 

As our study contains two classes of data which are imbalanced, the area under the curve (AUC) for ROC is not capable of fully reflecting model performance \cite{branco2016survey}. Figure \ref{fig:prec}(b) shows each  model's precision scores (the fraction of relevant articles among the retrieved articles) and recall scores (the fraction of the total amount of relevant articles that were actually retrieved) which better describe each model's performance under data imbalance. Model 1 attained a precision of 0.268 and a recall of 0.645. This suggests that the differences between the relevant documents and the irrelevant documents were not well expressed by the linear Boolean search criterion. 

Figure \ref{fig:prec}(b) shows Model 2 outperformed Model 1 at the same recall level. Model 3 with 15 clusters and the top 20 significant tokens improved AUC-PR from 0.34 to 0.5 compared to Model 2. Increasing the number of significant tokens to 50 and 100 improved AUC-PR to 0.56, while increasing to 250 tokens achieved a peak of 0.57. 

Under Model 3, a recall of 0.8 can be achieved while maintaining a precision score of 0.4. This reduces the amount of manual reading for our example by 80\%. Our example consists of about 10,000 total documents, about 1,000 of which are truly relevant, thus for Model 3 to correctly identify 800 relevant articles (80\% of 1,000 truly relevant cases), 800/0.4 = 2,000 documents will be labeled as potentially relevant. Researchers can then manually read the 2,000 labeled documents to identify the 800 relevant articles, reducing the amount of manual reading from 10,000 to 2,000. Other cut-off probabilities could be explored and would lead to different combination precision and recall scores as illustrated in Figure \ref{fig:prec}(b). In practice, people could rank the probability of relevance, $\hat{P}(Y_d = 1)$, and then prioritize manual reading of the highest probabilities. 



\begin{figure}[!htb]
  \centering
\begin{subfigure}[b]{0.45\textwidth}
  \centering
  \caption{}
  \includegraphics[width=3.2in]{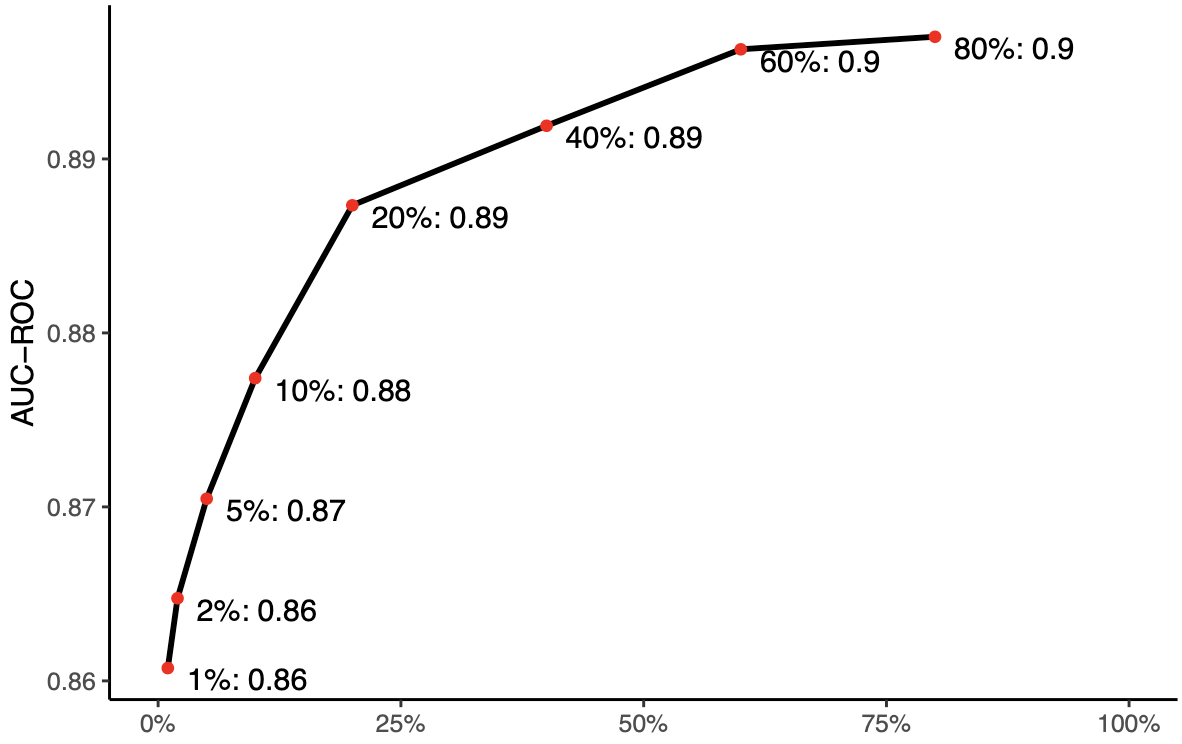}
\end{subfigure}
\begin{subfigure}[b]{0.45\textwidth}
  \centering
  \caption{}
  \includegraphics[width=3.2in]{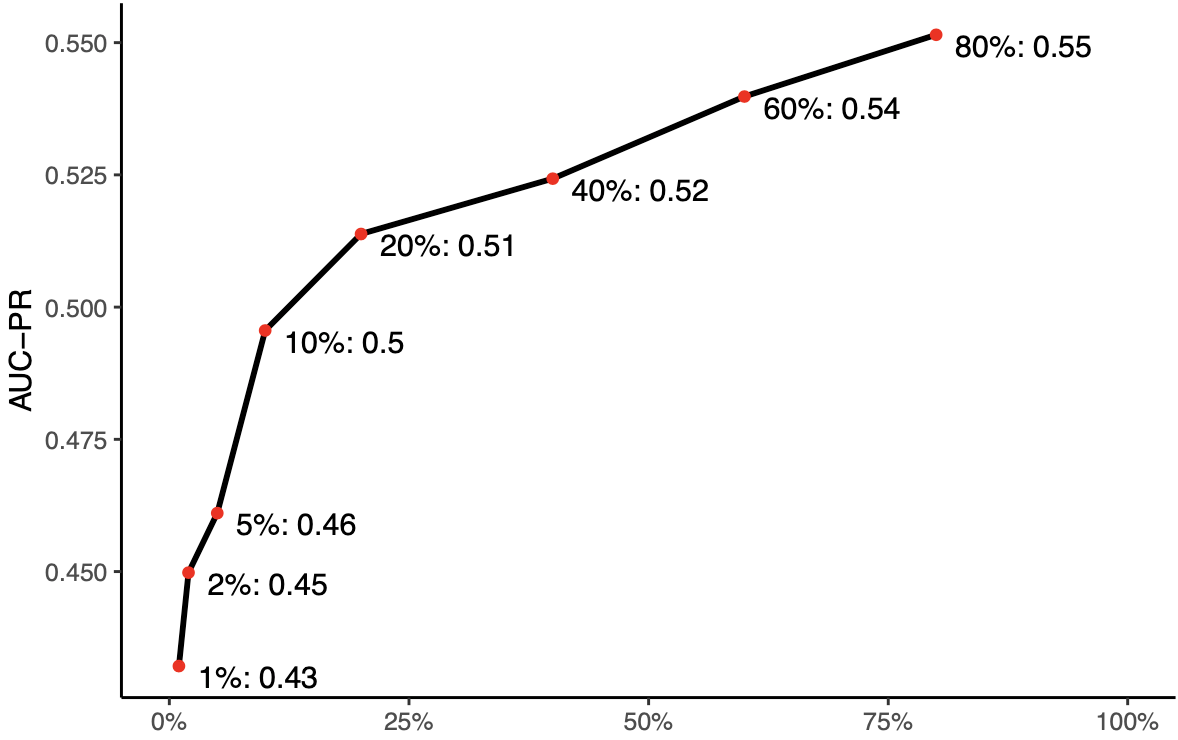} 
\end{subfigure}
\captionsetup{justification=centering,margin=2cm}
\caption{AUC for (a) ROC and (b) PR on testing data for Model 3 with different proportion of \\ pre-labeled documents from 1\% to 80\%}
\label{fig:precision-recall}
\end{figure}

The above 5-fold cross-validation results reflect the model performance when being trained on 80\% of the documents ($10,718$ in total). Next, we investigated the training data sample size that was needed to achieve a good model performance. We trained Model 3 with the top 250 significant tokens on different proportions of pre-labeled documents ($10,718$ in total) from 1\% up to 80\% and used cross validation to estimate the prediction accuracy on the test data. Figure \ref{fig:precision-recall} shows that we can attain an AUC for ROC above 80\% using only 1\% of pre-labeled documents (107 samples) as training data. As such, using Model 3 trained on just $107$ randomly selected training samples, we can still reduce the amount of manual reading for researchers by 68\% while identifying 80\% of truly relevant documents. The improvements for AUC-ROC and AUC-PR are both marginal when $2,144$ (20\% among $10,718$) or more pre-labeled documents were used. 




\section{Conclusion}\label{sec:conc}

The need to extract information through a systematic literature review arises in a wide range of domains. In this article, we proposed multiple models for automatically identifying documents that characterize key features of female sex workers. Empirical results showed that Model 3 outperformed the others using random forest on semantic clusters of key tokens and an additional set of tokens identified as significant. There are other machine learning algorithms available; we chose random forest for its ability to handle a large number of predictors and allow for high-order interaction effects.

Model 3 could be applied to identifying other populations most affected by HIV including clients of female sex workers, gay men and other men who have sex with men, people who use drugs, transgender populations, and incarcerated populations. We expect obtaining a reasonable classifier with taking 20\% of retrieved documents as training samples. The proposed classifier could also be used for more meaningfully assemble literature in other research areas; and for rapid documents screening with a tight schedule such as COVID related work during the crisis.

Finally, the trained classifiers can also be used as a sanity test, as mistakes are inevitable when manually screening thousands of documents. Any disagreements between manual and automatic classification would be addressed with an additional layer of review.
For instance, in some cases our model assigned high probabilities of relevant/irrelevant to documents manually labeled as the opposite. We provided 40 such documents to the manual screening team for verification: 18 documents labeled as irrelevant and 22 documents labeled as relevant. Among the 18 marked irrelevant documents, 5 of them are indeed relevant but missed during the manual labeling process; among the 22 marked relevant documents, 15 of them are identified as ``likely should not have been included". 



\section*{Acknowledgments}

This research was supported by National Institute of Allergy and Infectious Diseases of the National Institutes of Health under award number R01AI136664. Authors are grateful for helpful discussions with Amrita Rao, Jian Wu and Kunho Kim.



\subsection*{Financial disclosure}

None reported.

\subsection*{Conflict of interest}

The authors declare no potential conflict of interests.

\bibliography{wileyNJD-AMS}

\begin{thebibliography}{15}
\expandafter\ifx\csname natexlab\endcsname\relax\def\natexlab#1{#1}\fi
\expandafter\ifx\csname url\endcsname\relax
  \def\url#1{{\tt #1}}\fi
\expandafter\ifx\csname urlprefix\endcsname\relax\def\urlprefix{URL }\fi
\expandafter\ifx\csname doiprefix\endcsname\relax\def\doiprefix{doi:}\fi

\bibitem[{Agrawal et~al.(2000)Agrawal, Bayardo, and Srikant}]{agrawal2000}
Agrawal, R., R.~Bayardo, and R.~Srikant, 2000: Athena: Mining-based interactive
  management of text databases. {\it International Conference on Extending
  Database Technology\/}, Springer, 365--379.

\bibitem[{Angione(1975)}]{angione1975equivalence}
Angione, P.~V., 1975: On the equivalence of boolean and weighted searching
  based on the convertibility of query forms. {\it Journal of the American
  Society for Information Science (pre-1986)\/}, {\bf 26}, no. 2, 112.

\bibitem[{Branco et~al.(2016)Branco, Torgo, and Ribeiro}]{branco2016survey}
Branco, P., L.~Torgo, and R.~P. Ribeiro, 2016: A survey of predictive modeling
  on imbalanced domains. {\it ACM Computing Surveys (CSUR)\/}, {\bf 49}, no. 2,
  1--50.

\bibitem[{Collobert and Weston(2008)}]{collobert2008}
Collobert, R. and J.~Weston, 2008: A unified architecture for natural language
  processing: Deep neural networks with multitask learning. {\it Proceedings of
  the 25th international conference on Machine learning\/}, 160--167.

\bibitem[{Costelloe et~al.(2010)Costelloe, Metcalfe, Lovering, Mant, and
  Hay}]{costelloe2010effect}
Costelloe, C., C.~Metcalfe, A.~Lovering, D.~Mant, and A.~D. Hay, 2010: Effect
  of antibiotic prescribing in primary care on antimicrobial resistance in
  individual patients: systematic review and meta-analysis. {\it Bmj\/}, {\bf
  340}, c2096.

\bibitem[{Joachims(1998)}]{joachims1998}
Joachims, T., 1998: Text categorization with support vector machines: Learning
  with many relevant features. {\it European conference on machine learning\/},
  Springer, 137--142.

\bibitem[{Jones(1972)}]{jones1972statistical}
Jones, K.~S., 1972: A statistical interpretation of term specificity and its
  application in retrieval. {\it Journal of documentation\/}.

\bibitem[{Khan et~al.(2003)Khan, Kunz, Kleijnen, and Antes}]{khan2003five}
Khan, K.~S., R.~Kunz, J.~Kleijnen, and G.~Antes, 2003: Five steps to conducting
  a systematic review. {\it Journal of the royal society of medicine\/}, {\bf
  96}, no. 3, 118--121.

\bibitem[{Liaw and Wiener(2002)}]{randomforest}
Liaw, A. and M.~Wiener, 2002: Classification and regression by randomforest.
  {\it R News\/}, {\bf 2}, no. 3, 18--22.
\newline\urlprefix\url{https://CRAN.R-project.org/doc/Rnews/}

\bibitem[{Mechura(2016)}]{mechura2016data}
Mechura, M., 2016: Data structures in lexicography: from trees to graphs.,
  97--104.

\bibitem[{Moher et~al.(2009)Moher, Liberati, Tetzlaff, Altman, Group,
  et~al.}]{moher2009preferred}
Moher, D., A.~Liberati, J.~Tetzlaff, D.~G. Altman, P.~Group, et~al., 2009:
  Preferred reporting items for systematic reviews and meta-analyses: the
  prisma statement. {\it PLoS Med\/}, {\bf 6}, no. 7, e1000097.

\bibitem[{Norman et~al.(2012)Norman, Byambaa, De, Butchart, Scott, and
  Vos}]{norman2012long}
Norman, R.~E., M.~Byambaa, R.~De, A.~Butchart, J.~Scott, and T.~Vos, 2012: The
  long-term health consequences of child physical abuse, emotional abuse, and
  neglect: a systematic review and meta-analysis. {\it PLoS Med\/}, {\bf 9},
  no. 11, e1001349.

\bibitem[{Porter(2006)}]{porter2006algorithm}
Porter, M.~F., 2006: An algorithm for suffix stripping. {\it Program\/}.

\bibitem[{Schofield and Mimno(2016)}]{schofield2016comparing}
Schofield, A. and D.~Mimno, 2016: Comparing apples to apple: The effects of
  stemmers on topic models. {\it Transactions of the Association for
  Computational Linguistics\/}, {\bf 4}, 287--300.

\bibitem[{Whiteford et~al.(2013)Whiteford, Degenhardt, Rehm, Baxter, Ferrari,
  Erskine, Charlson, Norman, Flaxman, Johns, et~al.}]{whiteford2013global}
Whiteford, H.~A., L.~Degenhardt, J.~Rehm, A.~J. Baxter, A.~J. Ferrari, H.~E.
  Erskine, F.~J. Charlson, R.~E. Norman, A.~D. Flaxman, N.~Johns, et~al., 2013:
  Global burden of disease attributable to mental and substance use disorders:
  findings from the global burden of disease study 2010. {\it The lancet\/},
  {\bf 382}, no. 9904, 1575--1586.

\end{thebibliography}



\end{document}